%% file: ifacconf.tex
\documentclass{ifacconf}

\usepackage{graphicx}      
\usepackage{natbib}        

\usepackage{amsmath,amssymb,amsfonts}
\usepackage{algorithm}

\usepackage{algorithmic}
\usepackage{array}
\usepackage[caption=false,font=normalsize,labelfont=sf,textfont=sf]{subfig}
\usepackage{stfloats}
\usepackage{url}

\usepackage{enumerate}
\usepackage{color}
\usepackage{xcolor}
\usepackage{subfiles}
\usepackage{comment}
\usepackage{bm}

\newtheorem{theorem}{Theorem}

\newtheorem{lemma}{Lemma}
\newtheorem{remark}{Remark}

\newtheorem{proposition}{Proposition}
\newtheorem{definition}{Definition}

\newcommand{\mbr}{\mathbb{R}}
\newcommand{\transpose}{^\top}

\begin{document}
\begin{frontmatter}

\title{Feasible Force Set Shaping for a Payload-Carrying Platform Consisting of Tiltable Multiple UAVs Connected Via Passive Hinge Joints\thanksref{footnoteinfo}} 

\thanks[footnoteinfo]{*This work was supported by JSPS KAKENHI, Grant Number 21H01348, JST SPRING Grant Number JPMJSP2106, and the NSK Foundation for the Advancement of Mechatronics.}
\thanks{\copyright 2026 the authors. This work has been accepted to IFAC for publication under a Creative Commons Licence CC-BY-NC-ND.}

\author[First]{Takumi Ito} 
\author[Second]{Hayato Kawashima}
\author[Third]{Riku Funada} 
\author[Force]{Mitsuji Sampei}

\address[First]{National Institute of Advanced Industrial Science and Technology, Ibaraki, Japan (e-mail: takumi.ito@aist.go.jp)}
\address[Second]{Institute of Science Tokyo, Tokyo, Japan (e-mail: kawashima@sl.sc.e.titech.ac.jp).}
\address[Third]{Kyoto University, Kyoto, Japan (e-mail: funada@i.kyoto-u.ac.jp).}
\address[Force]{The Polytechnic University of Japan, Tokyo, Japan (e-mail: m-sampei@uitec.ac.jp).}

\begin{abstract}                
This paper presents a method for shaping the feasible force set of a payload-carrying platform composed of multiple Unmanned Aerial Vehicles (UAVs) and proposes a control law that leverages the advantages of this shaped force set. The UAVs are connected to the payload through passively rotatable hinge joints. The joint angles are controlled by the differential thrust produced by the rotors, while the total force generated by all the rotors is responsible for controlling the payload. The shape of the set of the total force depends on the tilt angles of the UAVs, which allows us to shape the feasible force set by adjusting these tilt angles. This paper aims to ensure that the feasible force set encompasses the required shape, enabling the platform to generate force redundantly \textemdash meaning in various directions. We then propose a control law that takes advantage of this redundancy.
\end{abstract}


\end{frontmatter}

\subfile{text/introduction}

\subfile{text/model}
\subfile{text/optimization}
\subfile{text/controller}
\subfile{text/simulation}

\section{CONCLUSION}
This work presented an analysis and control method for a payload-carrying platform utilizing multiple UAVs connected via passive hinge joints. First, we defined a feasible force set, which ensures local controllability based on the concept of hoverability, and named it a hoverable force set (HFS). The tilt of the UAVs allowed us to shape this HFS, enabling the platform to generate force in various directions. We developed a method for determining the tilt angles, which makes the HFS include the user-defined required force set (RFS).
Next, we introduced a new control strategy that takes advantage of the redundancy in feasible force generation while also ignoring the delay in tilt dynamics. Finally, simulations demonstrated the effectiveness of the proposed method, showing that it can effectively control the payload even in the presence of disturbances.

\bibliography{biblio}

\end{document}

%% file: text/introduction.tex
\section{INTRODUCTION}

The advancement of Unmanned Aerial Vehicles (UAVs) has enabled applications to be conducted automatically, such as agriculture (\cite{Tsouros2019agriculture}), environmental monitoring (\cite{10374239}), and inspection (\cite{jordan2018inspection}). Additionally, there is potential for using UAVs in payload transportation (\cite{Villa2020transportation}) due to increased package supplies and a labor shortage. Despite these diverse applications, conventional UAVs, consisting of multiple rotors pointing upward and placed on the same plane, are known as an underactuated system at $\mathrm{SE}(3)$ space (six-dimensional space). They need to cline themselves to move sideward, which is not suited for some tasks. Several approaches are being considered to make UAVs a fully actuated system in $\mathrm{SE}(3)$. 



The first approach is tilting and fixing the rotors of a UAV against its body. The theoretical analysis of the performance of UAVs with arbitrary structures is discussed in (\cite{8359200}). Additionally, many UAV designs have been developed based on concepts such as dynamic manipulability ellipsoids (\cite{8618982}) or the size of the feasible force set (\cite{case2024starshaped}). The tilt angle of UAVs in payload-carrying platforms with multiple UAVs has also been considered (\cite{Tan2018Heavy}). While this strategy simplifies the structure and model of UAVs, it results in a loss of thrust due to cancellation by internal forces.

\begin{figure}[!t]
    \centering
    \includegraphics[trim = 0 0 0 0, width=0.85\linewidth, clip]{"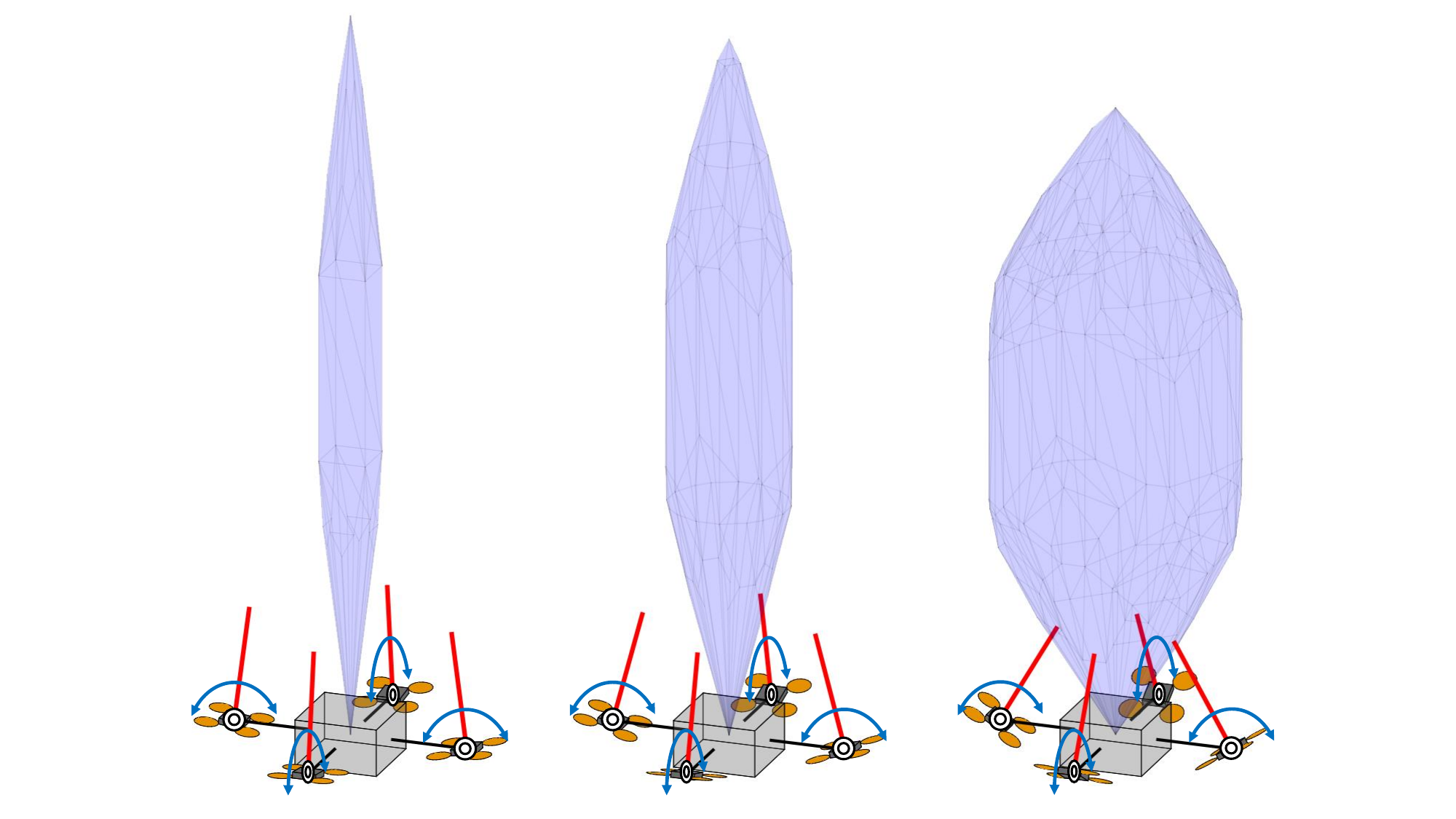"}
    \caption{Illustration of force sets feasible during hovering. Blue polytope, blue arrow, and red line are force set, tilt motion, and UAVs' direction, respectively. The force set is deformable by changing the tilt angles.}
    \label{fig:overview}
\end{figure}

The second approach involves adding extra actuators to tilt the rotors dynamically. In this method, a quadcopter can achieve full actuation by adjusting the thrust directions with servo motors (\cite{6225129}), and the same applies to hexacopters (\cite{8485627}). Additionally, a more complex structure with multiple links and servo motors can control a wider range of motion, including its own deformation (\cite{8258850}).
Using multiple UAVs for payload carrying has also been explored, incorporating servo motors to enhance maneuverability (\cite{Waner2018flightworthness}). However, as noted in some studies (\cite{10598313, IRIARTE2024104761}), this approach requires additional mechanisms and actuators, which can increase the overall weight of the UAV.

The final approach involves using passive tilting mechanisms controlled by the differential thrust of rotors. This method eliminates the need for actuators to deform structures that do not contribute to thrust generation. Additionally, it helps avoid undesirable dynamics coupling, such as the reaction torque of the actuators.
Various mechanisms have implemented passive tilting, including hinge joints (\cite{10598313, RUAN2023102927}), universal joints (\cite{IRIARTE2024104761}), gimbal joints (\cite{10214628}), and ball joints (\cite{Nguyen2018balljoint}). 

In this final approach, it is important to consider the delays of tilting dynamics, as highlighted in (\cite{10598313}). In all studies, tilt angles are adaptively addressed by employing hierarchical controllers or by compensating directly through a low-level controller (\cite{10598313}).
However, these methods cannot completely eliminate the effects of delay. This paper proposes a method that disregards the delays of tilting dynamics by leveraging the redundancy of the platform. The platform is inherently redundant since it can generate force in various directions without changing the tilt angles. The set of the generatable force forms a polytope, as shown in Fig.~\ref{fig:overview}). 
In our proposed method, the tilt angles are adjusted to ensure that the platform has a feasible force set, which includes the user-defined required shape. Then, the force input for controlling the platform is generated within the current force set without waiting for a delay.
By utilizing this redundancy, we can effectively disregard the delay of tilt dynamics. To the authors' knowledge, the redundancy of passive structure UAVs has not been leveraged to control and mitigate delays in tilt dynamics.

To investigate redundancy, we utilize a polytope force set (\cite{Debele2022amsfault, Zhang2021amsoptim}), which can exactly represent the feasible set without approximations. This contrasts with the manipulability ellipsoid (\cite{manipulability_ellipsoid}), which is commonly used to assess the motion performance of robots, including UAVs (\cite{8618982}). However, the polytope set requires more computational load. To address this issue, we compute all values offline and, during control, extract the necessary data from the pre-computed results.

The contributions of this research can be summarized as follows:
First, we provide a method for determining the tilt angles that compensate for the feasible force set to satisfy the required shape. Next, to address the delay in tilt dynamics of the UAVs, we propose a new control scheme that leverages redundancy of the feasible force set, which utilizes pre-computed tilt angles. Finally, simulations are conducted to validate the benefits of the proposed method.

%% file: text/model.tex
\section{MODELING}

\subsection{Configuration}

\begin{figure}[!t]
    \centering
    \includegraphics[trim = 100 0 100 0, width=0.7\linewidth, clip]{"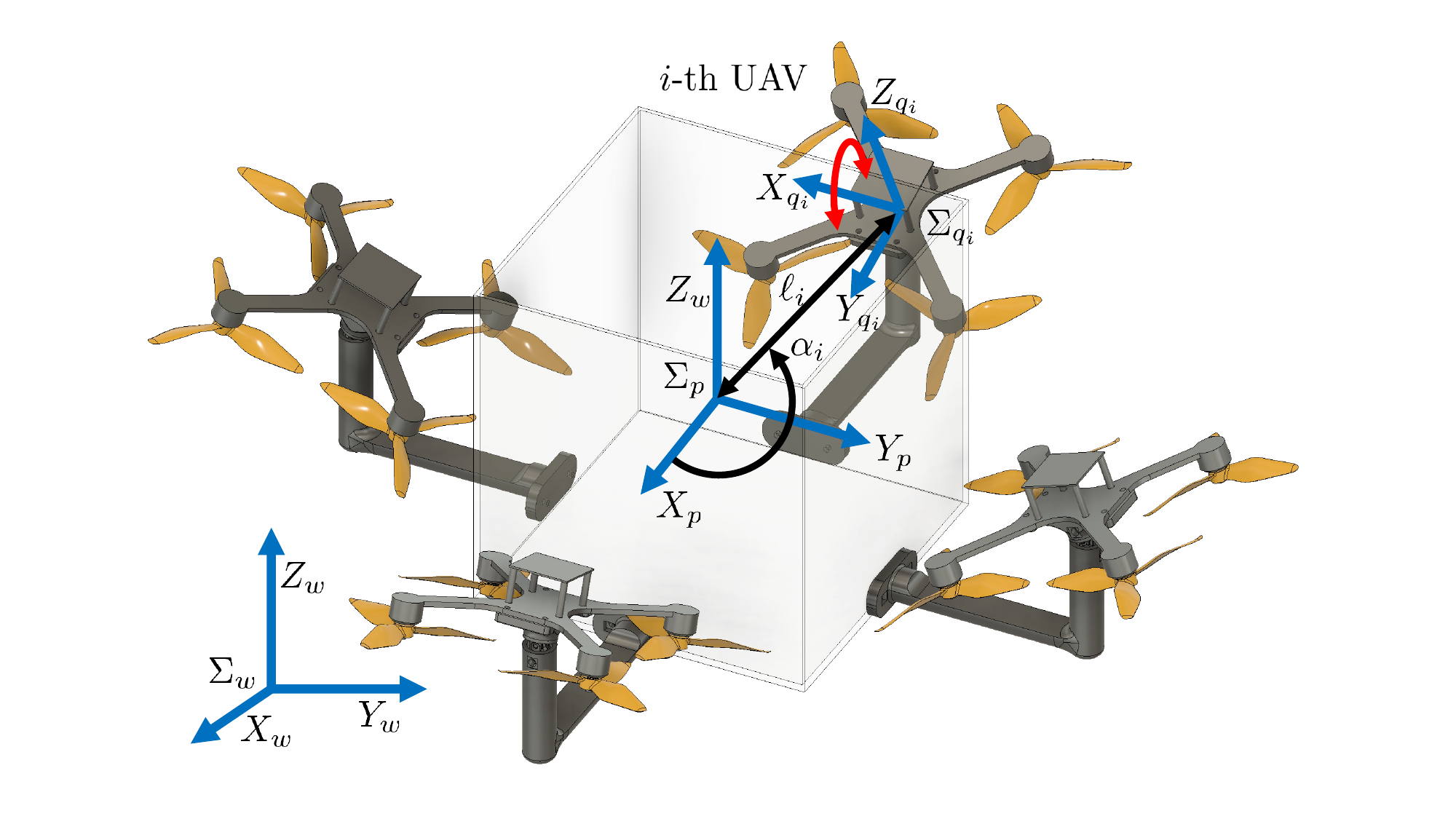"}
    \caption{An example of the payload-carrying platform with four UAVs.}
    \label{fig:platform}
\end{figure}

\begin{figure}[!t]
    \centering
    \includegraphics[trim = 100 0 100 0, width=0.7\linewidth, clip]{"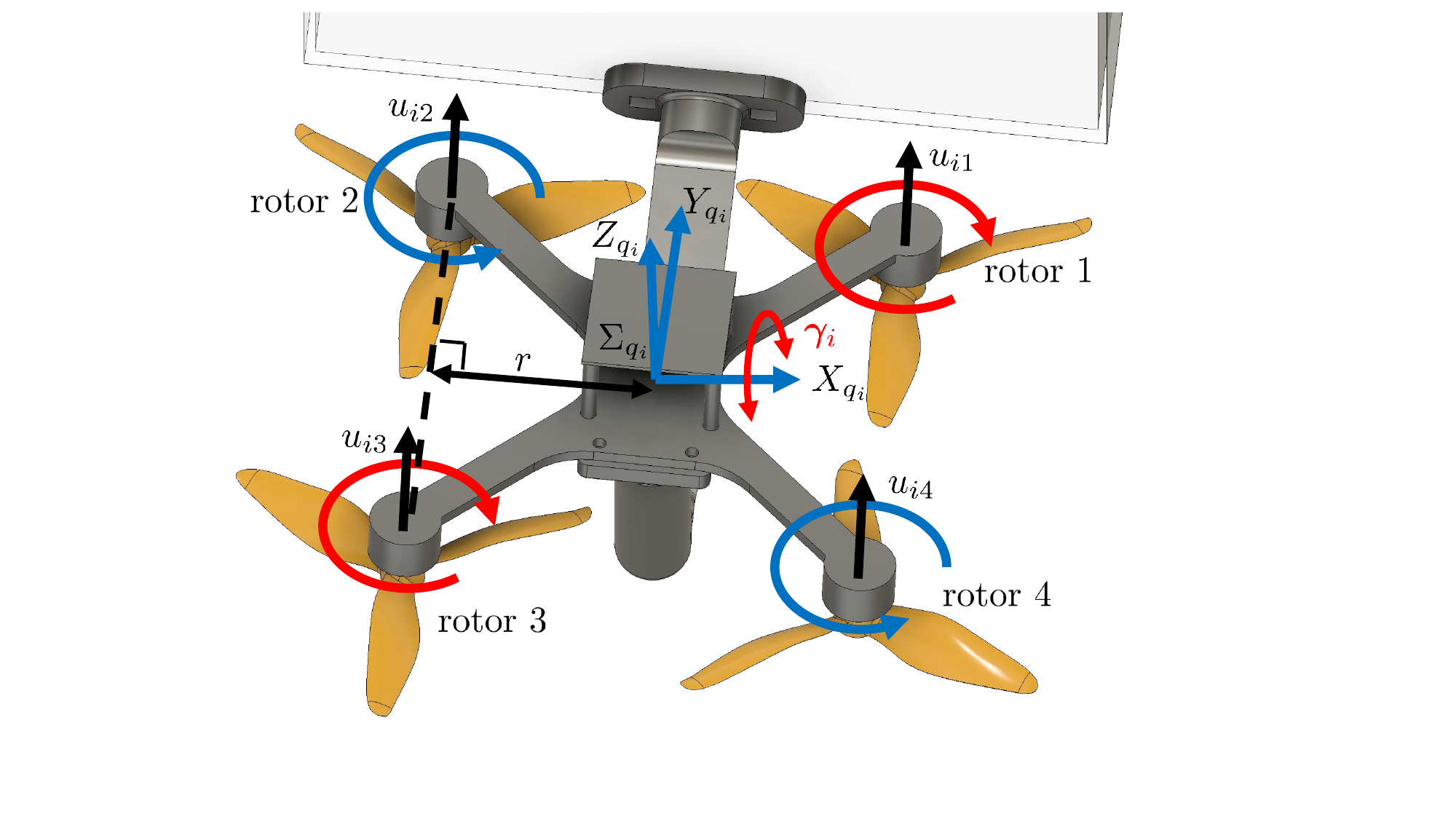"}
    \caption{Enlarged view of the UAV. One UAV has four rotors: two rotate clockwise, and the other two rotate counter-clockwise, arranged alternately at positions $[\pm r, \pm r, 0]^\top$ in $\Sigma_{q_i}$.}
    \label{fig:drone}
\end{figure}
This paper addresses a payload-carrying platform consisting of one payload and $N$ quadrotor UAVs. Fig.~\ref{fig:platform} illustrates an example with four UAVs. Each UAV is connected to the payload through a frictionless hinge joint, which allows the joint to rotate passively by the differential thrust of the rotors. The center of gravity (CoG) of each UAV is positioned at the rotational center of its joint. Additionally, the change in the moment of inertia of the platform due to the tilt of the UAVs is negligible, and we consider it to be constant.

There are some coordinate frames, as depicted in Fig.~\ref{fig:platform}. The world frame denoted as $\Sigma_w:~\{O_w,~X_w,~Y_w,~Z_w\}$ is a world-fixed frame whose origin is $O_w$, and $X_w$, $Y_w$, and $Z_w$ axes point east, north, and upward, respectively. The payload frame is denoted as $\Sigma_p:~\{O_p,~X_p,~Y_p,~Z_p\}$, and its origin $O_p$ is at CoG of the platform. The $i$-th UAV frame, $\Sigma_{q_i}:~\{O_{q_i},~X_{q_i},~Y_{q_i},~Z_{q_i}\}$, is defined for all UAVs whose origin $O_{q_i}$ is fixed to the CoG of the $i$-th UAV and is on the $X_p$-$Y_p$ plane and $X_{q_i}$ axis is orthogonal with line $O_p$-$O_{q_i}$. 
The $i$-th UAV tilts around the $X_{q_i}$ axis, and its angle is denoted as $\gamma_i$. When $\gamma_i=0$, the $Y_{q_i}$ points towards $O_p$ and $Z_{q_i}$ becomes parallel to $Z_p$. 
The length of the line $O_p$-$O_{q_i}$ is constant with $\ell_i$, and the angle between the line $O_p$-$O_{q_i}$ and the $X_p$ axis is denoted as $\alpha_i$. Suppose $\alpha_{i_1}\neq\alpha_{i_2}$ for all $i_1\neq i_2\in\{1,\cdots,N\}$.

\subsection{Thrust Transformation}
This subsection provides how the rotor thrusts of the UAVs work on the payload and joint motions. To simplify the discussion, we assume that all the rotors have the same properties and all the UAVs share the same structure, as shown in Fig.~\ref{fig:drone}. 
The $j$-th rotor of the $i$-th UAV generates the thrust in the rotational axis while generating the proportional counter torque in the opposite direction of rotation. Let the rotational velocity, the thrust, and the torque be $\omega_{ij}$, $u_{ij}$, and $\tau_{ij}$, respectively. Then, they are formulated as $u_{ij} = c_{u}\omega_{ij}^2,~ \tau_{ij} = \pm c_{\tau} \omega_{ij}^2$ with positive coefficients $c_{u}$ and $c_{\tau}$. The sign $\pm$ takes plus or minus when the rotor rotates clockwise or counter-clockwise, respectively. They can be rewritten as 
\begin{align}
    \tau_{ij} = \pm \kappa u_{ij},\quad \kappa = c_{\tau}/c_{u}. \label{eq:rotor_thrust}
\end{align}

Consider $u_{ij}$ as the control input, the wrench $\bm F_{q_i}$ that works on the $i$-th UAV in $\Sigma_{q_i}$ is written as a matrix product as 
\begin{align}
    \bm F_{q_i}
    = 
    \begin{bmatrix}
        \bm O_{2\times4} \\
    \begin{matrix}
        1 & 1 & 1 & 1 \\
        r & r & -r & -r \\
        -r & r & r & -r \\
        \kappa & -\kappa & \kappa & -\kappa
    \end{matrix}
    \end{bmatrix}
    \begin{bmatrix}
        u_{i1} \\ u_{i2} \\ u_{i3} \\ u_{i4}
    \end{bmatrix},\label{eq:thrust_trans_a}
\end{align}
where $r$ is described in Fig.~\ref{fig:drone}.

Since the UAV tilts around the $X_{q_i}$ axis, a torque to rotate joint angle is extracted from \eqref{eq:thrust_trans_a} as
\begin{align}
    \tau_{\gamma_i}= \begin{bmatrix}
        0 & 0 & 0 & 1 & 0 & 0 
    \end{bmatrix} \bm F_{q_i}. \label{eq:tau_betai}
\end{align}
The remaining elements are applied to the payload. Let the position and rotation of the $i$-th UAV frame $\Sigma_{q_i}$ in the payload frame $\Sigma_{p}$ be $\bm p_{pq_i}=\bm R_z(\alpha_i)[\ell_i~0~0]^\top\in\mbr^3$ and $\bm R_{pq_i}=\bm R_z(\alpha_i)\bm R_x(\gamma_i)\in\mathrm{SO}(3)$, respectively. Here, $\bm R_x$, $\bm R_y$, and $\bm R_z$ represents a rotation matrix around $x$, $y$, and $z$ axis, respectively. Then, the wrench generated by the $i$-th UAV expressed in $\Sigma_p$ is derived as
\begin{align}
    \bm F_{p_i} &= \underbrace{\begin{bmatrix}
                       \bm R_{pq_i}\transpose & \bm O_3\\
                       -\bm R_{pq_i}\transpose\widehat{\bm p}_{pq_i} & \bm R_{pq_i}\transpose
                   \end{bmatrix}}_{=:\mathrm{Ad}_{g_{pq_i}}\transpose}
                   \mathrm{diag}(1, 1, 1, 0, 1, 1) \bm F_{q_i},\label{eq:Fpi}
\end{align}
where $\mathrm{diag}()$ forms the diagonal matrix and $\bm O_3\in\mbr^{3\times 3}$ is zero matrix (similarly, $\bm O_{n\times m}\in\mbr^{n\times m}$ denotes zero matrix hereafter). The operator $~\widehat{}~$ provides $\widehat{\bm a}\bm b=\bm a\times\bm b$ for any vector $\bm a,~\bm b\in\mathbb{R}^3$. $\mathrm{Ad}_{g_{pq_i}}$ is a matrix called adjoint transformation, which is used for coordinate transformation of rigid bodies (\cite{hatanaka2015passivity}).

By combining \eqref{eq:tau_betai} for all the UAVs, we obtain a map from rotor thrusts to all joint torques as
\begin{align}
    \bm \tau_{\gamma}= \bm M_\gamma \bm u, \label{eq:M_gamma}
\end{align}
with 
$\bm u = \begin{bmatrix}
u_{11}\,\cdots\,u_{14} & & \cdots & & u_{N1}\,\cdots\,u_{N4} 
\end{bmatrix}^\top$ 
and 
$\bm \tau_\gamma = \begin{bmatrix}
\tau_{\gamma_1} & \cdots & \tau_{\gamma_N}
\end{bmatrix}^\top$.
Furthermore, by combining \eqref{eq:Fpi} for all the UAVs, we obtain
a map $[\bm M_{f}^\top~\bm M_{\tau}^\top]^\top$, which transforms rotor thrusts to the total wrench of the payload as
\begin{align}
    \begin{bmatrix}
        \bm f_p \\
        \bm \tau_p
    \end{bmatrix}
    =\sum_{i=1}^{N}\bm F_{p_i} = \begin{bmatrix}
                \bm M_{f} \\
                \bm M_{\tau}
            \end{bmatrix} \bm u.\label{eq:Mrp}
\end{align}
Finally, from \eqref{eq:M_gamma} and \eqref{eq:Mrp}, we obtain a map, which transforms rotor thrusts to the payload wrench and the joint torques as
\begin{align}
    \begin{bmatrix}
        \bm f_p \\
        \bm \tau_p \\
        \bm \tau_{\gamma}
    \end{bmatrix}
    = \underbrace{\begin{bmatrix}
        \bm M_{f} \\
        \bm M_{\tau} \\
        \bm M_\gamma
    \end{bmatrix}}_{=:\bm M_\text{all}} \bm u.\label{eq:Mall}
\end{align}

\begin{remark}
    Each UAV has four actuators and generates four degrees of freedom (DoF) wrench, namely upward thrust and three-dimensional torque in the UAV frame. While the torque around one axis controls the joint angle, the remaining three DoF, upward thrust, and two-dimensional torque, are used to control the payload. Hence, the total DoF controlling the payload is $3N$. The payload control can be redundant with more than or equal to three UAVs.
\end{remark}

\subsection{Equation of Motion}
Let the body velocity and the angular velocity in the payload frame be $\bm v$ and $\bm\omega$, respectively. The payload attitude is $\bm R_{wp}=\bm R_z(\psi)\bm R_y(\theta)\bm R_x(\phi)$ with $\phi$, $\theta$ and $\psi$, the roll-pitch-yaw angle. $m$ and $\bm J$ are the total mass and moment of inertia of the platform, respectively. $g$ is gravitational acceleration.
The equation of motion of the payload is formulated based on the Newton-Euler equation as 
\begin{align}
	\begin{bmatrix}
		m \bm I_3 & \bm O_3 \\
		  \bm O_3 &   \bm J
	\end{bmatrix}
	\begin{bmatrix}
		\dot{\bm v} \\
		\dot{\bm \omega}
	\end{bmatrix}
	+
	\begin{bmatrix}
		  \widehat{\bm \omega} m \bm v \\
		  \widehat{\bm \omega} \bm J \bm \omega
	\end{bmatrix}
	+
	\begin{bmatrix}
		m g \bm R_{wp}\transpose \bm e_3 \\
		  \bm 0_3
	\end{bmatrix}
	=
	  \begin{bmatrix}
            \bm f_p \\
            \bm \tau_p
        \end{bmatrix},
	\label{eq:NEeq}
\end{align}
where $\bm I_3\in\mbr^{3 \times 3}$ is identity matrix.
Equation \eqref{eq:NEeq} is defined in the payload frame $\Sigma_p$.
Additionally, the equation of motion of the joint rotation is formulated as
\begin{align}
    J_{x_i} \ddot{\gamma_i} &= \tau_{\gamma_i},\label{eq:gamma_modeli}
\end{align}
where $J_{x_i}$ is the moment of inertia around $X_{q_i}$ axis of the $i$-th UAV. Then, we get
\begin{align}
    \mathrm{diag}(J_{x_1},\cdots,J_{x_N}) \ddot{\bm \gamma} = \bm \tau_{\gamma},\label{eq:gamma_model}
\end{align}
with $\bm \gamma = [\gamma_1~\cdots~\gamma_N]^\top$.

%% file: text/optimization.tex
\section{OPTIMIZATION OF TILTING ANGLES}

\subsection{Feasible Force Set with Hoverability}
This subsection introduces a force set, which is attainable by the limitation of the rotor thrust inputs. 
Let the rotor input set be 
\begin{align}
    \mathcal{U}=\left\{\bm u \mid \bm 0 \leq \bm u \leq u_\text{max}\bm 1 \right\},\label{eq:U}
\end{align}
with the maximum thrust value $u_\text{max}$. $\bm 0$ and $\bm 1$ are vectors consisting of 0 and 1, respectively. In this paper, inequality $\leq$ of vectors is element-wise.

Although the arbitrary force in $\left\{\bm f_p \mid \bm f_p=\bm M_f\bm u,~\bm u\in\mathcal{U} \right\}$ can be generated by certain rotor thrusts, it is not guaranteed that the rotor thrusts keep hovering. As a remedy for this problem, we introduce the concept of hoverability (\cite{mochidahoverability}), the realizability of the static hovering of multirotor UAVs. 



\begin{lemma}[Hoverability conditions] \label{lem:hoverability}
    A multirotor UAV is hoverable if and only if the following two conditions are satisfied.
    \begin{enumerate}
        \item There exists input vector $\bm u_{eq}\in\mathcal{U}$ satisfying
            \begin{align} 
                \begin{bmatrix}
                    \bm M_{f} \\
                    \bm M_{\tau}
                \end{bmatrix} \bm u_{eq}
                = \begin{bmatrix}\bm f_p^\text{eq}\\ \bm O_{3\times 1} \end{bmatrix},\label{eq:hoverable1}
            \end{align}
            where $\bm f_p^\text{eq}$ is the force to keep the equilibrium state, e.g. $\bm f_p^\text{eq}=[0~0~mg]^\top$ under no disturbance situation.
        \item The matrix $[\bm M_{f}^\top~\bm M_{\tau}^\top]^\top$ satisfies
            \begin{align}
                \mathrm{rank}\left(\begin{bmatrix}
                    \bm M_{f} \\
                    \bm M_{\tau}
                \end{bmatrix}\right) \geq 4.
            \end{align}
    \end{enumerate}
    \vspace{0em}
\end{lemma}
This hoverability ensures the existence of the equilibrium state and input, as well as local controllability around the equilibrium state.
For a detailed explanation of hoverability, see the authors' work (\cite{mochidahoverability}).

\begin{figure}[!t]
    \centering
    \subfloat[]{\includegraphics[width=0.4\linewidth]{"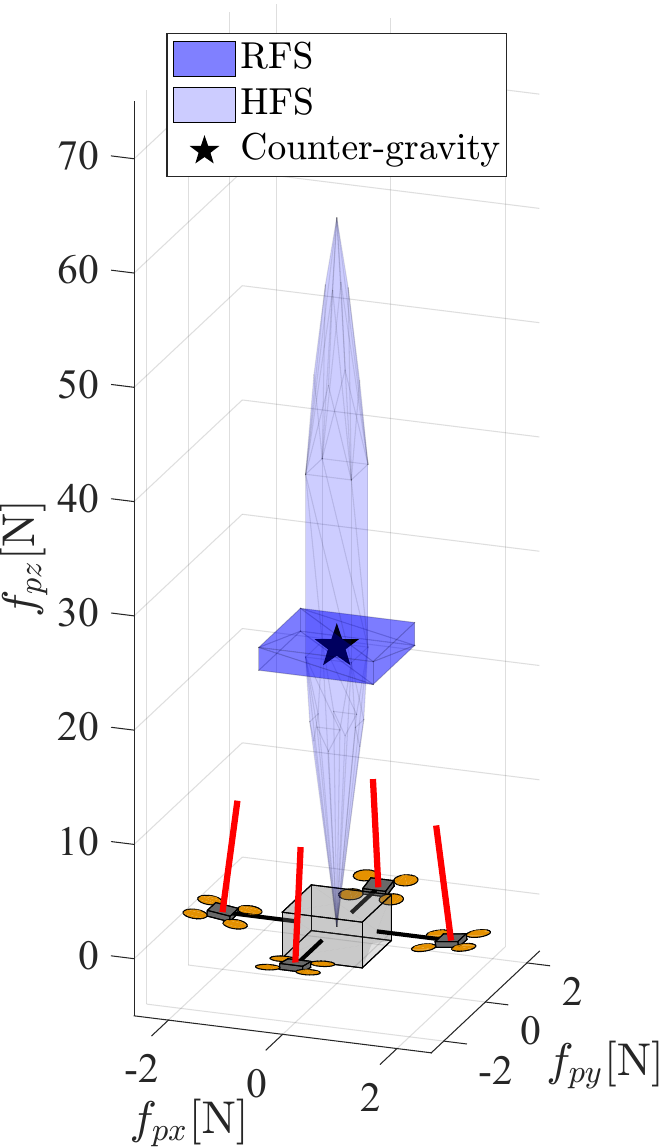"}%
    \label{fig:hfs24}}
    \hfil
    \subfloat[]{\includegraphics[width=0.4\linewidth]{"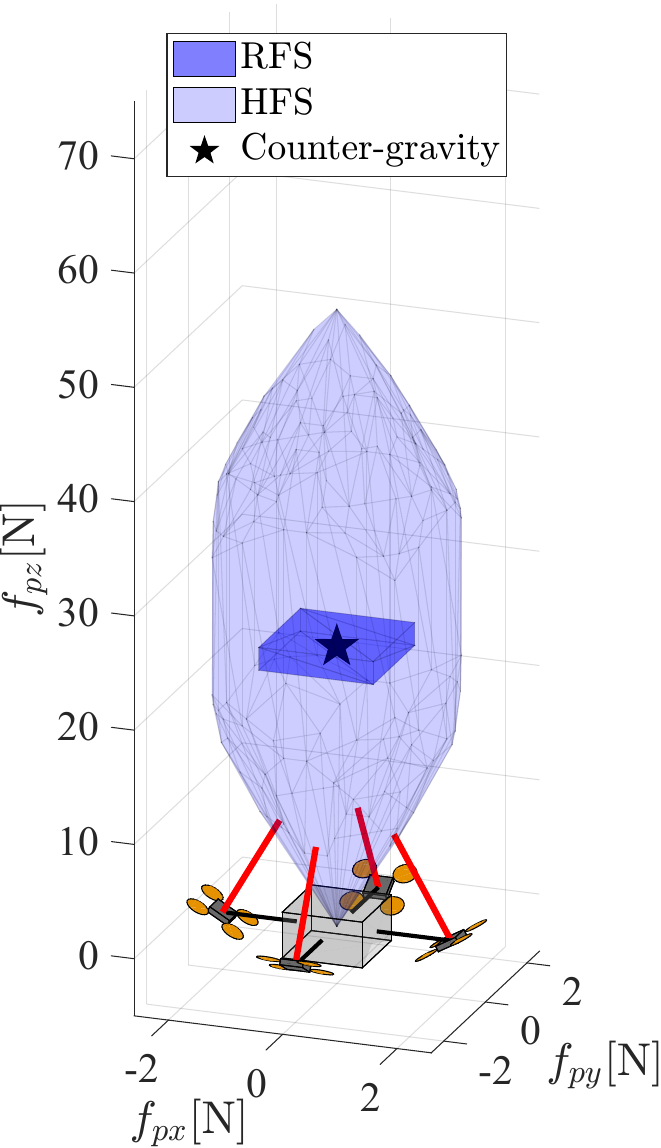"}%
    \label{fig:hfs6}}
    \caption{Examples of HFS and RFS. HFS and RFS are drawn as light blue and deep blue polytopes, respectively. $x$ and $y$ axes of force are enlarged five times of the $z$ axis. (a) $\gamma_i=-\pi/24$, (b) $\gamma_i=-\pi/6,~\forall i\in\{1,\cdots,N\}$.}
    \label{fig:hfs}
\end{figure}

To examine the hoverability of the platform, let us define a force set as follows.
\begin{definition}[Hoverability force set (HFS)]\label{def:HFS}
    A hoverability force set (HFS) is a set of forces that is attainable during the payload hovering. 
    \begin{align}
        \mathcal{F}_H =
        \left\{\bm f_p~\middle|~\begin{bmatrix}\bm f_p \\ \bm 0 \end{bmatrix}=\begin{bmatrix}\bm M_f \\ \bm M_\tau \end{bmatrix} \bm u,~\bm u\in\mathcal{U} \right\}.\label{eq:HFS}
    \end{align}
    \vspace{0em}
\end{definition}
Since $\bm M_f$ and $\bm M_\tau$ depend on the tilt angle $\bm \gamma$, the HFS also depends on $\bm \gamma$. We can shape the HFS by controlling the joint tilt angles, as shown in Fig.~\ref{fig:hfs}. Note that $x$ and $y$ axes of the force plot are enlarged for better visibility.
Then, the following theorem examines the hoverability of the platform.
\begin{theorem}[Hoverability condition of the platform] \label{thm:hove_platform}
    The platform is hoverable if and only if the HFS includes the equilibrium force $\bm f_p^\text{eq}$.
\end{theorem}
\begin{pf}
    The platform is hoverable if and only if it satisfies Lemma~\ref{lem:hoverability}.
    The statement of Theorem~\ref{thm:hove_platform} is equal to the condition Lemma~\ref{lem:hoverability}-1).
    Then, we have to check the condition Lemma~\ref{lem:hoverability}-2).
    Let $S_a$ and $C_a$ be aberrations of $\sin(a)$ and $\cos(a)$, respectively.
    From \eqref{eq:thrust_trans_a}, \eqref{eq:Fpi}, and \eqref{eq:Mrp}
    , matrix $[\bm M_{f}^\top~\bm M_{\tau}^\top]^\top$ can be rewritten as
    \begin{align}
        \begin{bmatrix}\bm M_f \\ \bm M_\tau \end{bmatrix} 
        = \tilde{\bm M}
        \left(\bm I_N \otimes \begin{bmatrix}
            1 & 1 & 1 & 1 \\
            r & r & -r & -r \\
            -r & r & r & -r \\
            \kappa & -\kappa & \kappa &-\kappa
        \end{bmatrix}\right),\label{eq:Mrp_real}
    \end{align}
    with 
    \begin{align}
        \tilde{\bm M} =& \begin{bmatrix}
            \begin{matrix}
                0 \\
                S_{\gamma_1} \\
                C_{\gamma_1}
            \end{matrix} & \bm O_3 & \cdots & \begin{matrix}
                0 \\
                S_{\gamma_N} \\
                C_{\gamma_N}
            \end{matrix} & \bm O_3 \\
            \ast & \begin{matrix}
                0 & S_{\alpha_1} & 0 \\
                0 & C_{\alpha_1}C_{\gamma_1} & S_{\gamma_1} \\
                0 & -C_{\alpha_1}S_{\gamma_1} & C_{\gamma_1}
            \end{matrix} & \cdots & \ast & \begin{matrix}
                0 & S_{\alpha_N} & 0 \\
                0 & C_{\alpha_N}C_{\gamma_N} & S_{\gamma_N} \\
                0 & -C_{\alpha_N}S_{\gamma_N} & C_{\gamma_N}
            \end{matrix}  
        \end{bmatrix}.\label{eq:Mtilde}
    \end{align}
    where $\otimes$ is Kronecker product. Then, the rank of $[\bm M_f^\top~\bm M_\tau^\top]^\top$ is equal to the rank of $\tilde{\bm M}$ because the matrix within the round bracket in \eqref{eq:Mrp_real} is nonsingular.
    Then, by rearranging the columns of \eqref{eq:Mtilde}, we obtain 
    \begin{align}
        &\mathrm{rank}(\tilde{\bm M}) \notag\\
        &=\! \mathrm{rank}\left(\left[
            \begin{matrix}
                \begin{matrix}
                    0 \\
                    S_{\gamma_1} \\
                    C_{\gamma_1}
                \end{matrix} \cdots
                \begin{matrix}
                    0 \\
                    S_{\gamma_N} \\
                    C_{\gamma_N}
                \end{matrix} 
                 & \bm O_{3\times N} \\
                 \ast & \bm O_{3\times N} 
            \end{matrix}
            \right.\right. \notag\\ 
            &\hspace{5.5em}\left.\left.
            \begin{matrix}
            \bm O_{3\times N} & \bm O_{3\times N} \\
            \begin{matrix}
                S_{\alpha_1}\\
                C_{\alpha_1}C_{\gamma_1} \\
                -C_{\alpha_1}S_{\gamma_1}
            \end{matrix}  \cdots 
            \begin{matrix}
                S_{\alpha_N}\\
                C_{\alpha_N}C_{\gamma_N} \\
                -C_{\alpha_N}S_{\gamma_N}
            \end{matrix}
            &
            \begin{matrix}
                0 \\
                S_{\gamma_1} \\
                C_{\gamma_1}
            \end{matrix}  \cdots
            \begin{matrix}
                0 \\
                S_{\gamma_N} \\
                C_{\gamma_N}
            \end{matrix}
            \end{matrix}
            \right]\right)\notag\\
        &=\! \mathrm{rank}\left(\left[
            \begin{matrix}
                0 \\
                S_{\gamma_1} \\
                C_{\gamma_1}
            \end{matrix} \cdots
            \begin{matrix}
                0 \\
                S_{\gamma_N} \\
                C_{\gamma_N}
            \end{matrix} \right]\right) \notag\\ 
            &\quad +
            \mathrm{rank}\left(\left[
            \begin{matrix}
            \begin{matrix}
                S_{\alpha_1}\\
                C_{\alpha_1}C_{\gamma_1} \\
                -C_{\alpha_1}S_{\gamma_1}
            \end{matrix}  \cdots 
            \begin{matrix}
                S_{\alpha_N}\\
                C_{\alpha_N}C_{\gamma_N} \\
                -C_{\alpha_N}S_{\gamma_N}
            \end{matrix}
            &
            \begin{matrix}
                0 \\
                S_{\gamma_1} \\
                C_{\gamma_1}
            \end{matrix}  \cdots
            \begin{matrix}
                0 \\
                S_{\gamma_N} \\
                C_{\gamma_N}
            \end{matrix}
            \end{matrix}
            \right]\right).\label{eq:Mrank}
    \end{align}
    
    Let $i_1\neq i_2\in\{1,\cdots,N\}$ be two different UAVs' indices. 
    The determinant of the matrix, obtained by extruding columns $i_1$, $i_2$, and $N+i_1$ from the second term of \eqref{eq:Mrank}, is nonzero if 
    \begin{align}
        \left|\begin{matrix}
            S_{\alpha_{i_1}} & S_{\alpha_{i_2}} & 0 \\
            C_{\alpha_{i_1}}C_{\gamma_{i_1}} & C_{\alpha_{i_2}}C_{\gamma_{i_2}} & S_{\gamma_{i_1}}\\
            -C_{\alpha_{i_1}}S_{\gamma_{i_1}} & -C_{\alpha_{i_2}}S_{\gamma_{i_2}} & C_{\gamma_{i_1}}
        \end{matrix}\right| \neq 0.
    \end{align}
    This is equivalent to the following condition to hold.
    \begin{align}
        \cos(\gamma_{i_1}-\gamma_{i_2})\neq \frac{\tan(\alpha_{i_2})}{\tan(\alpha_{i_1})}.\label{eq:detM1}
    \end{align}
    The same procedure with columns $i_1$, $i_2$, and $N+i_2$ gives
    \begin{align}
        \cos(\gamma_{i_1}-\gamma_{i_2})\neq \frac{\tan(\alpha_{i_1})}{\tan(\alpha_{i_2})}.\label{eq:detM2}
    \end{align}
    Because $\alpha_{i_1} \neq \alpha_{i_2}$ and $\frac{\tan(\alpha_{i_1})}{\tan(\alpha_{i_2})}>1$ or $\frac{\tan(\alpha_{i_1})}{\tan(\alpha_{i_2})}>1$, either \eqref{eq:detM1} or \eqref{eq:detM2} is always true. 
    Hence, at least a set of three columns guarantees the second term of \eqref{eq:Mrank} being 3.
    Since the first term of \eqref{eq:Mrank} is more than or equal to 1, $\mathrm{rank}(\tilde{\bm M})\geq 4$ holds. This means the platform always satisfies the condition Lemma~\ref{lem:hoverability}-2).
\end{pf}

\subsection{Tilt Angle Optimization}
As mentioned above, HFS is deformable by tilting the UAVs. This subsection provides the method to decide tilt angles once the required force set is given.

We define the required force set (RFS) as a convex polytope set as
\begin{align}
    \mathcal{F}_R &= \mathrm{conv}(\mathcal{V}_R),\label{eq:RFS}
\end{align}
where $\mathcal{V}_R=\{\bm v_1,~\bm v_2,\cdots\}$ is a vertex set of RFS and $\mathrm{conv}()$ provides a convex hull. The number of vertices is up to the shape of RFS, e.g., eight for a cuboid and much more number for a complex shape. As an example of RFS, a rectangular cuboid is drawn in Fig.~\ref{fig:hfs}.
From a practical viewpoint, we can design the center force point as a nominal force input used to track a trajectory and the volume as an acceptable variation of the force input due to feedback.

The platform is regarded as achieving the requirement when the HFS completely includes the RFS. For example, the platform in Fig.~\ref{fig:hfs}~\subref{fig:hfs24} does not achieve the requirement, in contrast to Fig.~\ref{fig:hfs}~\subref{fig:hfs6}.
Although Fig.~\ref{fig:hfs}~\subref{fig:hfs6} achieves the requirement, it is energy-inefficient because the UAVs tilt at large angles. The rotors need to generate excess thrust, part of which is canceled as an internal force in the net force. 

We propose the following optimization problem to find tilt angles that allow the HFS to achieve the requirement while keeping them small angles.
\begin{align}
    \bm\gamma^\ast = 
    \underset{\bm \gamma}{\mathrm{arg\,min}} \quad  & -\left|\mathcal{V}_R \cap \mathcal{F}_H\right| + \frac{\|\bm\gamma\|_2^2}{N\gamma_\text{max}^2+\epsilon}\label{eq:evaluate} \\
    \mathrm{s.t.}\quad & |\gamma_i|\leq\gamma_\text{max},~\forall i\in\{1,\cdots, 4\}, \notag
\end{align}
where $\epsilon$ is a small positive value, and $\gamma_\text{max}$ is a positive value. $|\mathcal{A}|$, for any set $\mathcal{A}$, returns the number of elements.
The first term maximizes the number of the vertices of the RFS in the HFS (minimizing the minus value) and takes an integer value. This term is used to detect the inclusion between the RFS and the HFS. A detailed calculation is provided in the following subsection.
The second term minimizes the $L_2$ norm of the tilt angles. This term suppresses large tilt angles.
\begin{remark}\label{rmk:multimodal}
    Because the objective function \eqref{eq:evaluate} is non-differentiable and multimodal, we employ the particle swarm optimization (PSO) algorithm offline, which searches in the global domain without the gradient of the objective function.
\end{remark}

\subsection{Detection of Inclusion}

This subsection provides a way to detect the inclusion between the RFS and HFS. As both are convex sets, the HFS completely includes the RFS when all the vertices of the RFS are internal to the HFS. 
The minimization of \eqref{eq:evaluate} does not obviously guarantee the inclusion of the RFS in the HFS because \eqref{eq:evaluate} is a multi-objective optimization. 
Then, the following theorem provides the conditions under which the optimal solution satisfies the inclusion.
\begin{theorem}[Inclusion and optimality]\label{thm:inclusion}
    Let $N_{R}$ be the number of the vertices of the RFS. 
    Then, the RFS is completely included in the HFS if the optimal solution of \eqref{eq:evaluate} is smaller than $-N_R+1$.
\end{theorem}
\begin{pf}
    The second term of \eqref{eq:evaluate} is 
    \begin{align}
        \frac{\|\bm\gamma\|_2^2}{N\gamma_\text{max}^2+\epsilon} &< \frac{\|\bm\gamma\|_2^2}{N\gamma_\text{max}^2} = \frac{1}{N}\sum_{i=1}^{N}\frac{\gamma_i^2}{\gamma_\text{max}^2}\notag\\
        &\leq \frac{1}{N}\sum_{i=1}^{N}1 = 1.
    \end{align}
    Then, the following inequality holds.
    \begin{align}
         -\left|\mathcal{V}_R \cap \mathcal{F}_H\right| + \frac{\|\bm\gamma\|_2^2}{N\gamma_\text{max}^2+\epsilon}< -\left|\mathcal{V}_R \cap \mathcal{F}_H\right| + 1.\label{eq:objfunccond}
    \end{align}
    Hence, if the optimal value is smaller than $-N_R+1$. It satisfies that $\left|\mathcal{V}_R \cap \mathcal{F}_H\right|=N_R$.
\end{pf}

Next, we provide a method to examine whether each vertex of the RFS is included in the HFS or not.
From Definition~\ref{def:HFS}, a vertex $\bm v_k\in\mathcal{V}_R$ is in the HFS if there exists a rotor input $\bm u$ that satisfies
\begin{align}
    \begin{bmatrix}\bm v_k \\ \bm 0 \end{bmatrix} 
    =\begin{bmatrix}\bm M_f \\ \bm M_\tau \end{bmatrix} \bm u, \quad \exists\bm u\in\mathcal{U}.\label{eq:FinH}
\end{align}
Note that the rotor input $\bm u$ satisfying \eqref{eq:FinH} is not unique because the inverse mapping of \eqref{eq:FinH} is underdetermined.
Therefore, $\bm v_k$ is an internal point of the HFS if there exists at least one combination of rotor thrusts that satisfies \eqref{eq:FinH}.
To find satisfying rotor thrusts, we provide the following problem.
\begin{align}
    \mathrm{minimize}\qquad & \left\|\tilde{\bm u}\right\|_\infty \label{eq:fr_Linfty}\\
    \mathrm{s.t.}\qquad & \tilde{\bm u} = \frac{2}{u_\text{max}}\left(\bm u-\frac{u_\text{max}}{2}\bm 1\right),\label{eq:utilde}\\
                               & \begin{bmatrix} \bm v_k \\ \bm 0 \end{bmatrix}                  
                               = \begin{bmatrix} \bm M_f \\ \bm M_\tau \end{bmatrix}\bm u,\notag
\end{align}
where $\|~~\|_\infty$ calculates $L_\infty$ norm.
This is a problem of minimizing $L_\infty$ norm and can be rewritten to linear programming. Then, the inclusion is examined by the following proposition.
\begin{proposition}\label{prop:innerp}
    The vertex $\bm v_k$ is included in the HFS if and only if the solution of the \eqref{eq:fr_Linfty} is less than or equal to 1.
\end{proposition}
\begin{pf}
    If $\|\tilde{\bm u}\|_\infty\leq 1$, then $-\bm 1\leq\tilde{\bm u}\leq \bm 1$.
    From \eqref{eq:utilde}, we obtain$\bm u = u_\text{max}/2\left(\tilde{\bm u} + \bm 1\right)$.
    Hence, $0\leq \|\bm u\|_\infty\leq u_\text{max}$.
    Then, the optimal value $\bm u$ is in the rotor input set $\mathcal{U}$ and satisfies \eqref{eq:FinH}.
\end{pf}

Finally, by solving \eqref{eq:fr_Linfty} for all vertices, we can get the value of the first term of \eqref{eq:evaluate}. 
The whole calculation process to determine the tilt angles is shown in Algorithm \ref{alg:optim_shematic}. 

\begin{figure}[t]
    \begin{algorithm}[H]
        \caption{Overview of the optimization problem}
        \label{alg:optim_shematic}
        \begin{algorithmic}[1]
            \REQUIRE $\mathcal{V}_R$
            \ENSURE  $\bm\gamma^\ast$
            \WHILE{tolerances and stopping criteria}
                \STATE $\bm\gamma^\text{cand} \gets$ candidate tilt angles (PSO)
                \STATE $\mathit{count}=0$
                \FOR{$k=1$ to $|\mathcal{V}_R|$}
                    \STATE solve \eqref{eq:fr_Linfty} with $\bm\gamma^\text{cand}$ and vertex $\bm v_k\in\mathcal{V}_R$
                    \IF{$\|\tilde{\bm u}\|_\infty \leq 1$}
                        \STATE $\mathit{count} \gets \mathit{count}+1$
                    \ENDIF
                \ENDFOR
                \STATE evaluate \eqref{eq:evaluate} with $\left|\mathcal{V}_R \cap \mathcal{F}_H\right|=\mathit{count}$
            \ENDWHILE
            \STATE\RETURN $\bm\gamma^\ast=\bm\gamma^\text{cand}$
        \end{algorithmic}
    \end{algorithm}
\end{figure}


\begin{table}[t]
    \centering
    \caption{Physical parameters. $i\in\{1,\cdots,4\}$.}
    \label{tab:phisical_params}
    \begin{tabular}{l|r|l|r}
        \hline\hline
        $m$ &  2.5 kg & $\bm J$ & $\mathrm{diag}$(0.05,0.05,0.05) [kgm${}^2$]\\
        $J_{x_i}$ & 0.005 kgm${}^2$ & $\kappa$ & 0.011 \\
        $\ell_i$ & 0.22 m & $\alpha_i$ & $(i-1)\pi/2$ [rad]\\
        $r$ & 0.08 m  & $f_\text{max}$ & 4 N\\
        \hline
    \end{tabular}
\end{table}

\begin{figure}[t]
    \centering
    \includegraphics[trim = 0 0 0 0, width=\linewidth, clip]{"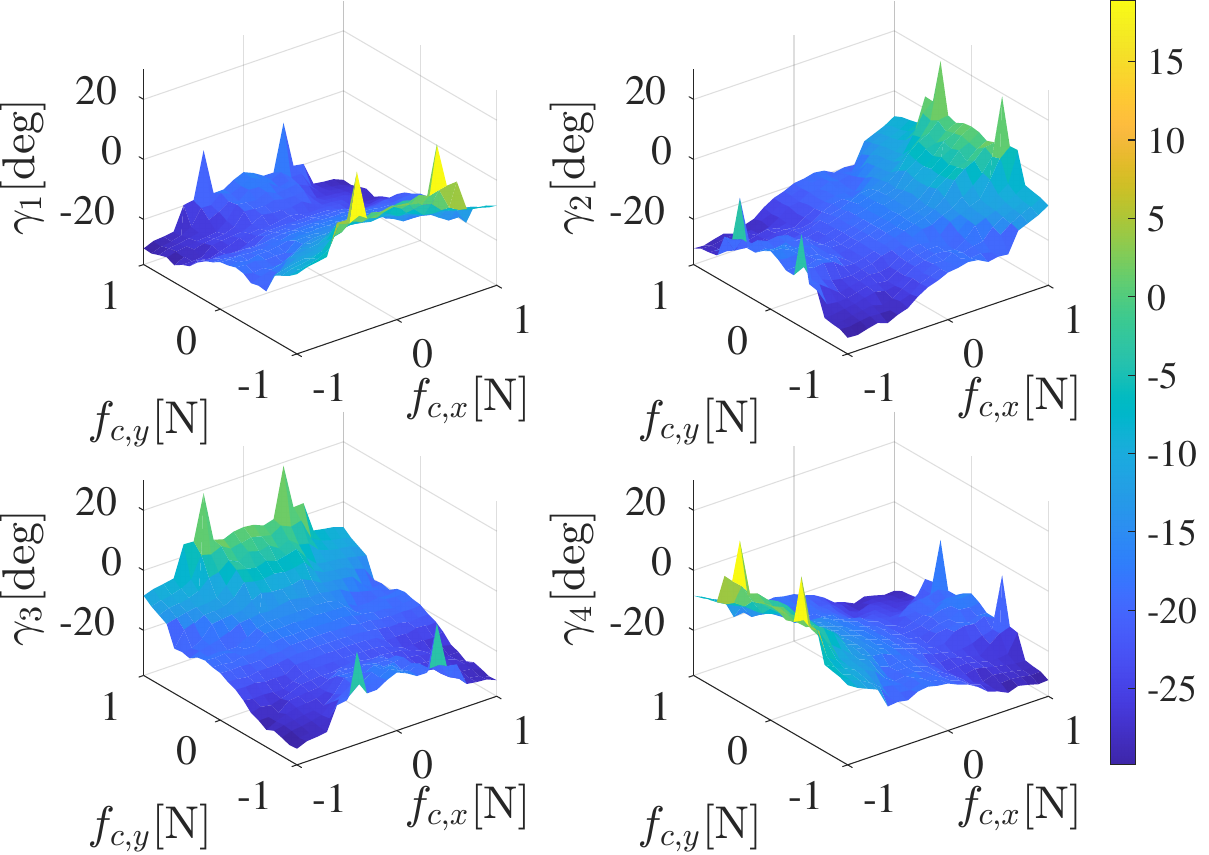"}
    \caption{The optimized tilt angles.}
    \label{fig:opt_angles_table}
\end{figure}

\begin{figure}[t]
    \centering
    \includegraphics[trim = 0 0 0 0, width=\linewidth, clip]{"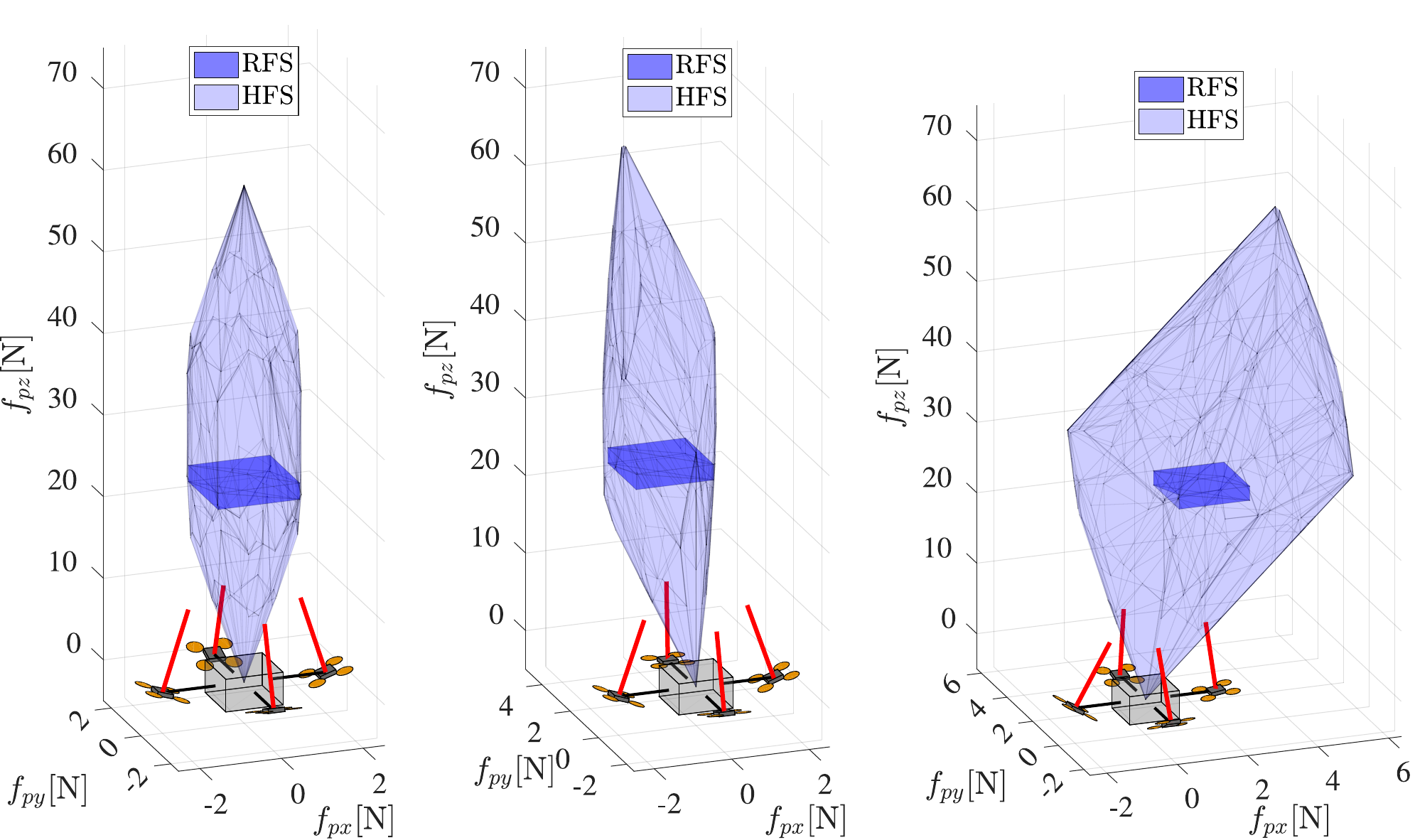"}
    \caption{HFSs shaped by optimized tilt angles to satisfy the RFSs. The geometric centers of the RFSs are $[0.0~0.0~mg]^\top$, $[0.0~1.0~mg]^\top$, and $[1.0~1.0~mg]^\top$, respectively, from the left to right figure.}
    \label{fig:exmpl_poly}
\end{figure}

\subsection{Examples}\label{sss:example}
We provide several examples to illustrate the proposed method, which will be used in the upcoming simulation. Table \ref{tab:phisical_params} displays the physical parameters utilized for computing models and optimization problems.
We define the vertices of RFS as having a cuboid shape as
\begin{align}
    \mathcal{V}_R=&\{f_{c,x}-1,~f_{c,x}+1\}\times\{f_{c,y}-1,~f_{c,y}+1\}\notag\\
    &\times\{f_{c,z}-1,~f_{c,z}+1\},
\end{align}
where $\times$ is Cartesian product and $[f_{c,x}~f_{c,y}~f_{c,z}]^\top$ is a geometric center of RFS. Then, the optimization problem \eqref{eq:evaluate} is solved with $\forall f_{c,x},\forall f_{c,y}\in\{0.0,~0.1,~\cdots,~1.0\}$ while $f_{c,z}=mg$ is fixed.
Then, the optimized tilt angles of the RFSs with the range $\forall f_{c,x},\forall f_{c,y}\in\{-1.0,~-0.9,~\cdots,~1.0\}$ are decided using the symmetry of the platform.
Calculating optimal angles using a polytope set takes much time, so these angles are computed offline for later use in the controller.

The optimized tilt angles are shown in Fig.~\ref{fig:opt_angles_table}. Several examples of HFS and RFS are displayed in Fig.~\ref{fig:exmpl_poly}.
All optimal solutions satisfy Theorem~\ref {thm:inclusion}. They indicate that the proposed method can determine tilt angles that make HFS include RFS while keeping the tilt angles small.

\begin{remark}
    Setting a large RFS enables better disturbance rejection and ignores larger delays in tilt dynamics. However, this could lead to large tilt angles, resulting in thrust waste in the net force. This is energy insufficient; users can design RFS while considering this tradeoff.
\end{remark}


%% file: text/controller.tex
\begin{figure*}[!t]
    \centering
    \includegraphics[trim = 0 0 0 0, width=0.8\linewidth, clip]{"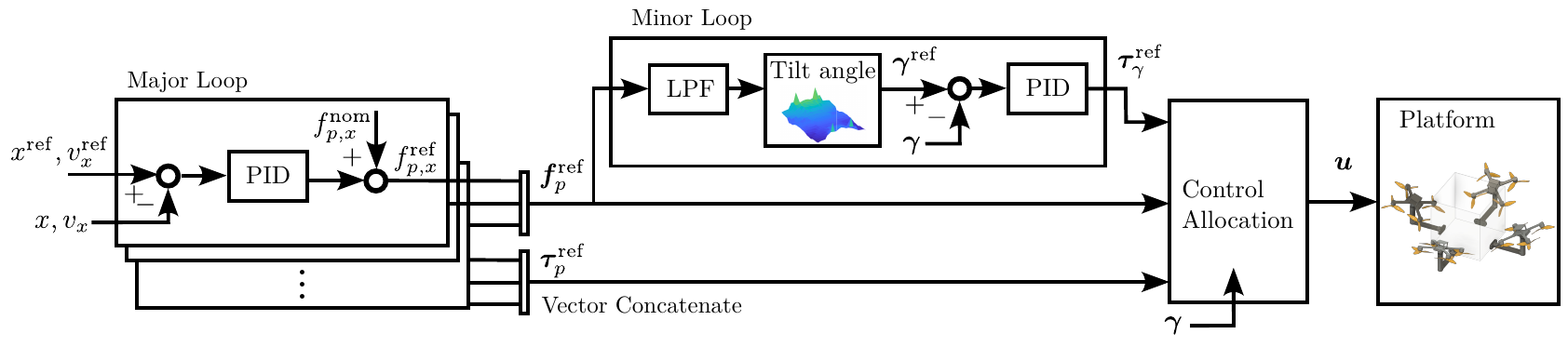"}
    \caption{Controller.}
    \label{fig:controller}
\end{figure*}

\section{CONTROLLER}
This section provides the controller that utilizes the redundancy of HFS to address the delay of the tilt dynamics.
Fig. \ref{fig:controller} shows the whole controller configuration.

The major loop controls the position $x,~y,~z\in\mbr$ and orientation of the payload. They generate the references $\bm f_p^\text{ref}=[f_{p,x}^\text{ref}~f_{p,y}^\text{ref}~f_{p,z}^\text{ref}]^\top$ and $\bm \tau_p^\text{ref} =[\tau_{p,x}^\text{ref}~\tau_{p,y}^\text{ref}~\tau_{p,z}^\text{ref}]^\top$ to control $x,~y,~z,~\phi,~\theta$, and $\psi$, respectively, by using PID controllers and the nominal forces $f_{p,x}^\text{nom}$, $f_{p,y}^\text{nom}$, $f_{p,z}^\text{nom}$, and $\tau_{p,x}^\text{nom}=\tau_{p,y}^\text{nom}=\tau_{p,z}^\text{nom}=0$.

The minor loop controls the tilt angles based on the reference forces computed by the major loop. Once the reference force is provided, the minor loop first uses a low-pass filter (LPF) to prevent aggressive changes in the reference force, which helps prevent aggressive changes in the reference of the tilt angles. Then, the references of the tilt angles $\bm\gamma^\text{ref}$ are decided using the pre-optimized tilt angles shown in Fig~\ref{fig:opt_angles_table} with $f_{c,x}=f_{p,x}^\text{ref}$ and $f_{c,y}=f_{p,y}^\text{ref}$. Since the tilt angles are computed with discretized values of $f_{c,x}$ and $f_{c,y}$, the tilt angle reference is generated by using cubic interpolation. 
The LPF causes a delay in the generation of the reference of tilt angles, but we can disregard it, as discussed later in Remark \ref{rmk:controller}.
The minor loop finally generates reference torque $\bm \tau_\gamma^\text{ref}$ to control tilt angles by PID controllers.

Once the payload wrench and the joint torque references are obtained, we have to allocate them to rotor thrusts, i.e., inverse mapping of \eqref{eq:Mall}. The solution of inverse mapping is not unique because the mapping \eqref{eq:Mall} is underdetermined.
While we focus on the feasibility in the optimization part, in control, we would like to allocate the reference wrench evenly to all rotors. Hence, we solve the optimization
\begin{align}
    \bm u^\ast = 
    \underset{\bm u}{\mathrm{arg\,min}}\quad & \mathrm{max}(\bm u)-\mathrm{min}(\bm u) \label{eq:allocation}, \\
    \mathrm{s.t.}\quad
                & \begin{bmatrix}
                    \bm f_p^{\text{ref}} \\
                    \bm \tau_p^{\text{ref}} \\
                    \bm \tau_{\gamma}^{\text{ref}}
                \end{bmatrix}
                = \bm M_\text{all} \bm u,~ \bm u \in \mathcal{U}. \notag
\end{align}
The optimization problem~\eqref{eq:allocation} finds rotor thrusts that minimize the difference between the maximum and minimum thrust. Moreover, \eqref{eq:allocation} can be rewritten to linear programming, which allows us to solve efficiently. 
\begin{remark}\label{rmk:controller}
    While the tilt angle reference is decided based on the force reference and controlled by the minor loop, the control allocation \eqref{eq:allocation} can be computed with current tilt angles. Since the HFS is shaped to have the required volume, the current reference force is attainable if the current HFS includes it, even if the tilt angles are delayed because of its dynamics and the LPF.
\end{remark}

%% file: text/simulation.tex
\section{SIMULATION}

\subsection{Simulation Settings}
We utilize a platform consisting of four UAVs for our simulation. All physical parameters are listed in Table \ref{tab:phisical_params}, and PID gains are set as $(1, 0.1, 1)$, $(10, 10, 10)$, and $(20, 1, 5)$, for translation, rotation, and tilt angles controllers, respectively. 
As LPF, we use a first-order system with a time constant of 1 second, represented by a transfer function: $1/(s+1)$.

\begin{figure}[!t]
    \centering
    \includegraphics[trim = 0 0 100 0, width=\linewidth, clip]{"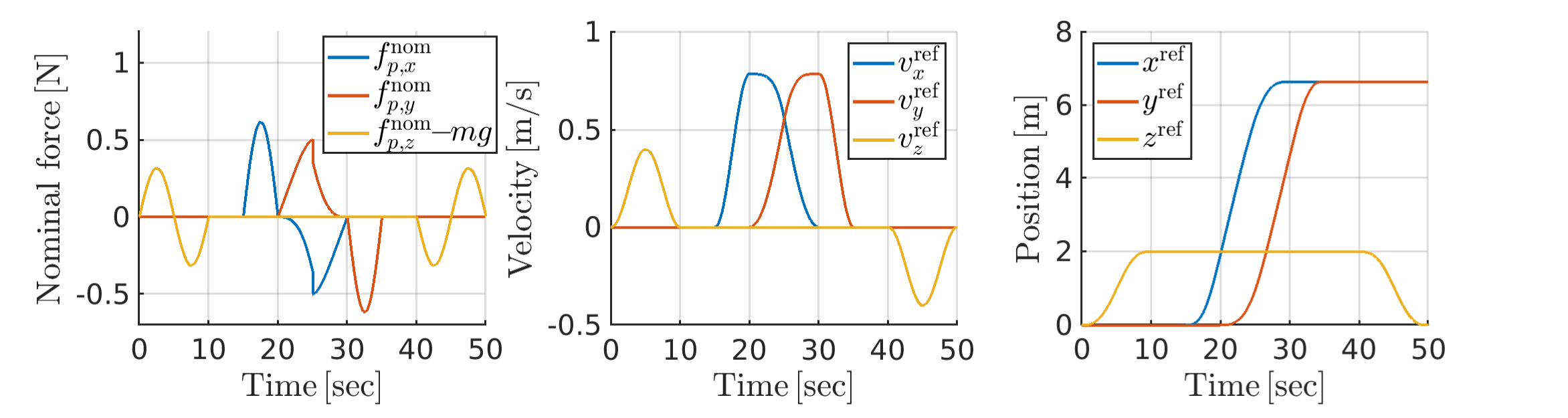"}
    \caption{Nominal force and trajectory of the simulation.}
    \label{fig:trajectory}
\end{figure}

Fig. \ref{fig:trajectory} shows the reference trajectory of the position and translational velocity of the payload in the world frame and the nominal payload force used to track it. This trajectory is designed to ensure continuous acceleration by employing trigonometric functions and clothoid curves. The trajectory for the payload orientation and angular velocity is maintained at zero throughout the simulation to keep the payload horizontal.

In the simulation environment, there is a wind disturbance of $0.5$ N directed positively along the $Y_w$ axis within the region defined by $1\leq x\leq4$, as illustrated as a gray area in Fig. \ref{fig:result}~\subref{fig:result_2D}. We assume that the platform is capable of detecting wind disturbances with its sensors and adding them to nominal forces.

The tilt angles are optimized with the same setting of subsection \ref{sss:example}. Then, we can utilize the optimized tilt angles depicted in Fig.~\ref{fig:opt_angles_table}. We assume the reference forces $f_{p,x}^\text{ref}$ and $f_{p,y}^\text{ref}$ do not exceed the range of Fig.~\ref{fig:opt_angles_table}, namely $f_{p,x}^\text{ref},~f_{p,y}^\text{ref}\in[-1.0, 1.0]$. Note that this assumption can be easily relaxed by computing optimal tilt angles for a wider range of RFSs.

\subsection{Simulation Results}
\begin{figure}[!t]
  \centering
  \subfloat[]{\includegraphics[trim = 50 0 50 0, width=0.49\linewidth, clip]{"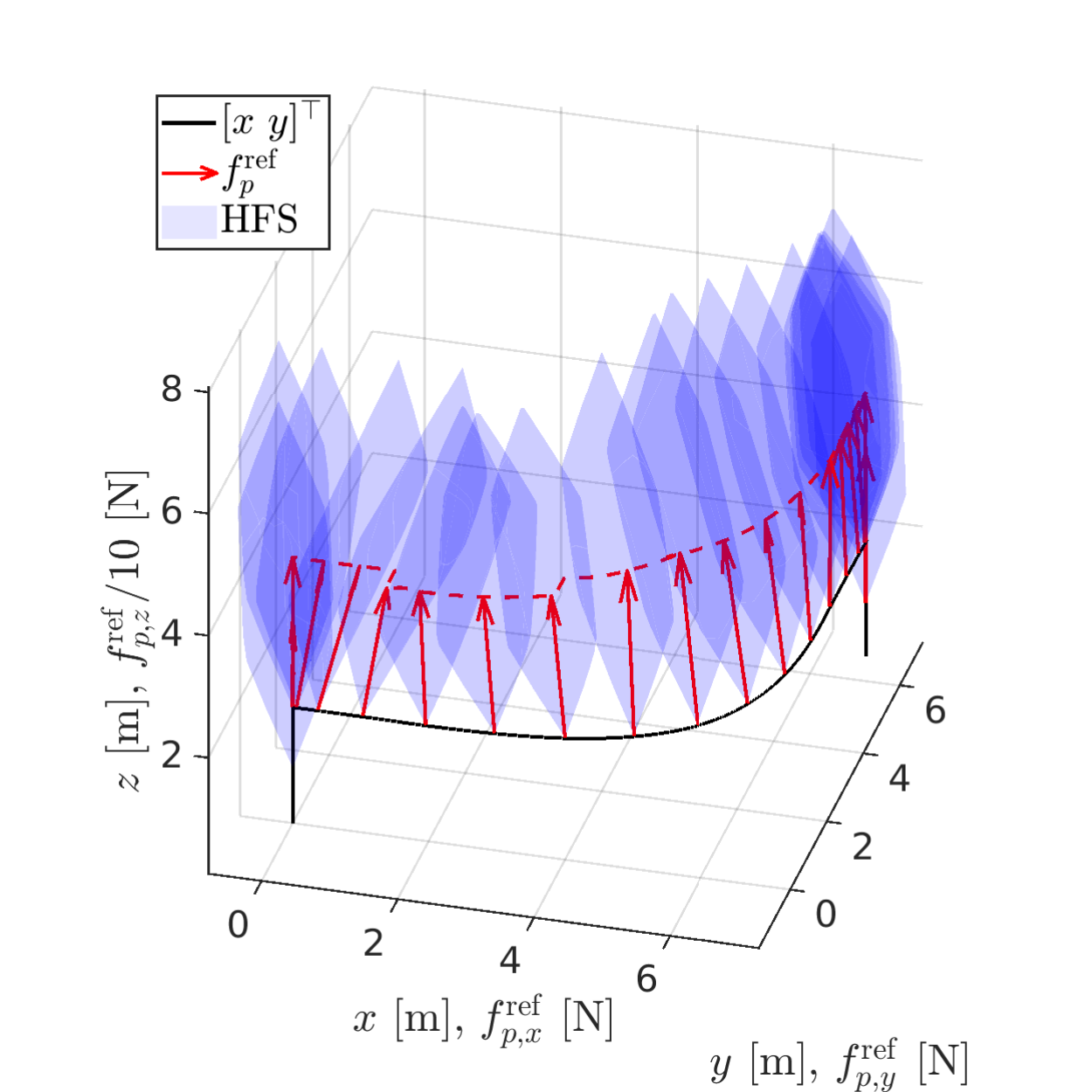"}%
  \label{fig:result_3D}}
  \hfil
  \subfloat[]{\includegraphics[trim = 0 0 0 0, width=0.49\linewidth, clip]{"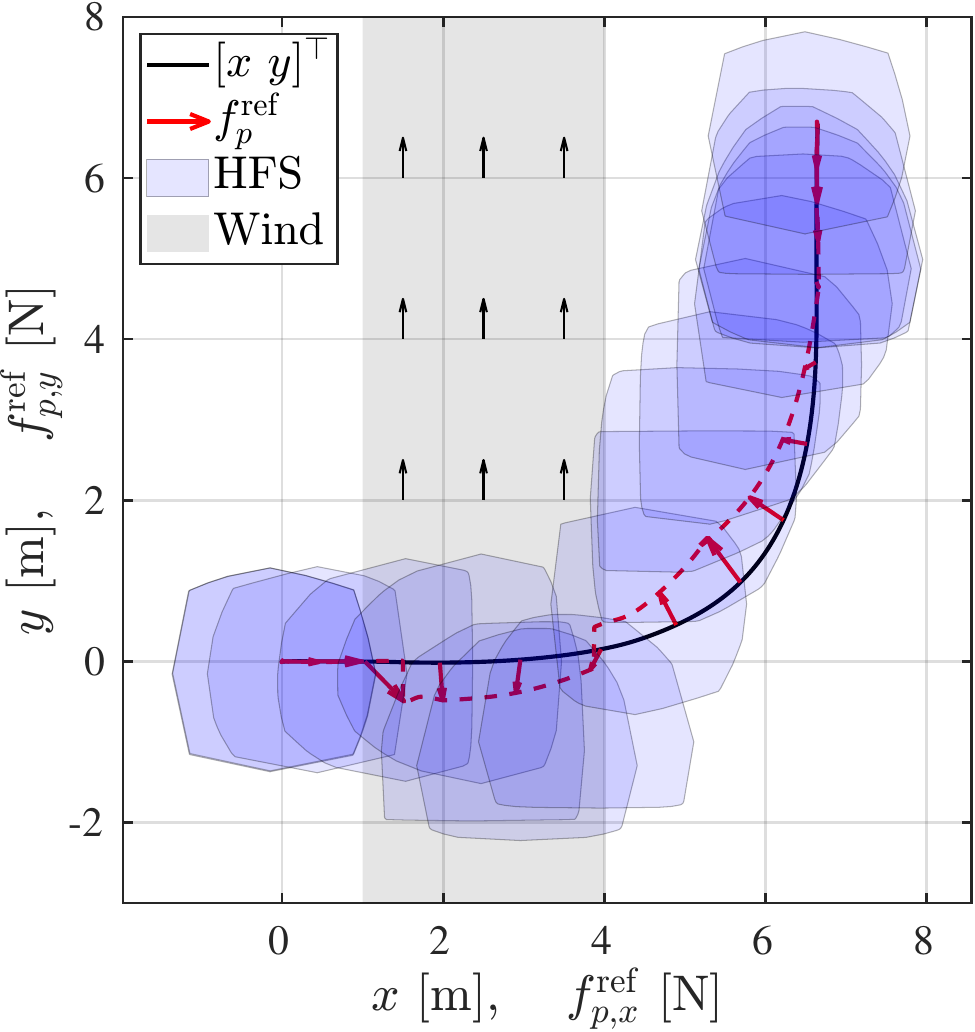"}%
  \label{fig:result_2D}}
  \caption{Result of the simulation. (a) 3D view. (b) 2D top view. The platform is not displayed for better visibility. The evolution of the position is plotted with the black line. The force reference is the red dot line. The HFSs and the reference force vectors (snapshots with an interval of 1.25 seconds) are shown as blue polytopes and red arrows, with the origin at the payload position. The $z$ element of force is displayed with a 1/10 scale for ease of viewing. The wind disturbance zone is shown with a gray area only in the 2D view.}
  \label{fig:result}
\end{figure}

\begin{figure}[!t]
  \centering
  \subfloat[]{\includegraphics[height=6.7em]{"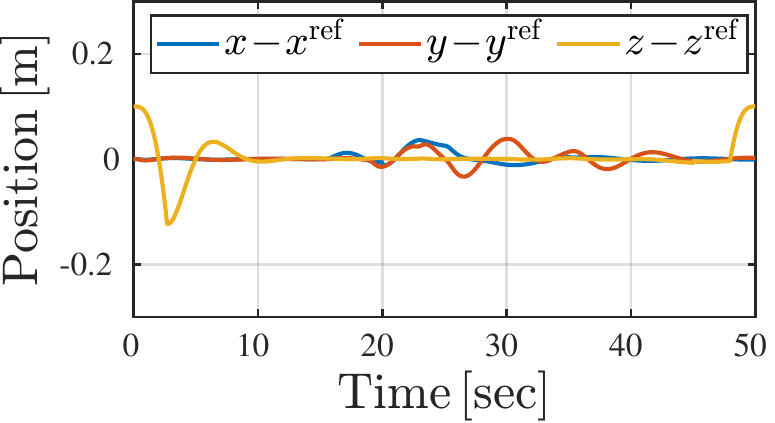"}%
  \label{fig:result_pos}}
  \hfil
  \subfloat[]{\includegraphics[height=6.7em]{"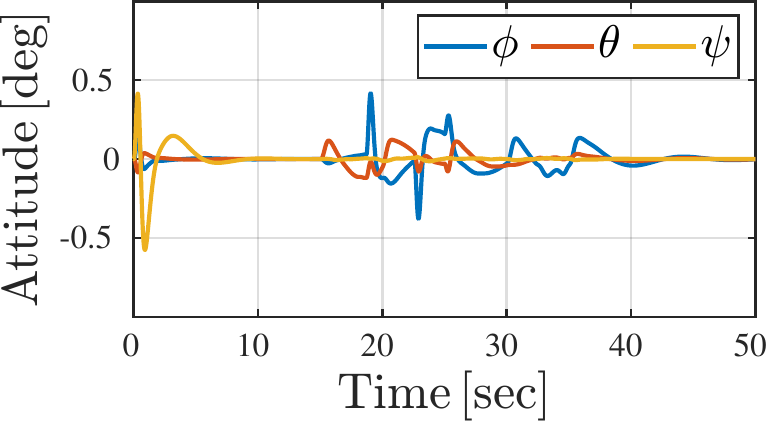"}%
  \label{fig:result_ori}}
  \caption{Trajectory tracking error. (a) Position error. (b) Orientation error.}
  \label{fig:result_error}
\end{figure}

\begin{figure}[!t]
  \centering
  \subfloat[]{\includegraphics[height=7em]{"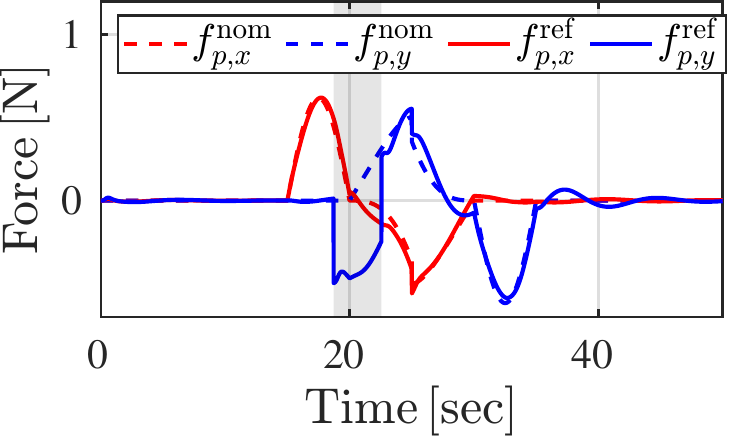"}%
  \label{fig:result_force}}
  \hfil
  \subfloat[]{\includegraphics[height=7em]{"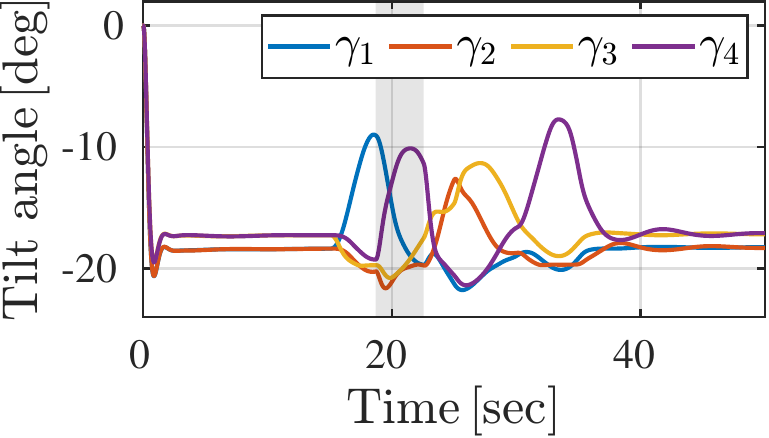"}%
  \label{fig:result_angle}}
  \caption{Evolution of the reference force and tilt angles. (a) Reference force with nominal force. (b) Tilt angles. The gray area displays the time when the payload is encountering the wind disturbance.}
  \label{fig:result2}
\end{figure}

Fig.~\ref{fig:result} shows the result of the simulation with 3D and 2D top views. Fig.~\ref{fig:result_error} shows the evolution of the tracking error of position and orientation. Fig.~\ref{fig:result2} shows the evolution of reference forces and tilt angles. 
The error of position from the trajectory value is shifted within 0.11 m. The error of orientation is shifted within 0.6 deg. 

The HFS is designed to include the volume of the RFS, which allows the reference force to shift within its range, as shown in Fig.~\ref{fig:result}. This occurs despite the delays associated with the HFS, including delays in tilt angle dynamics and the delays in generating the reference tilt angle due to the LPF.

The plots in the gray area of Fig. \ref{fig:result}~\subref{fig:result_2D} and Fig. \ref{fig:result2}~\subref{fig:result_force} demonstrate that the proposed control strategy effectively handles disturbances by leveraging the redundancy of the HFS. Since the HFS has a volume of RFS and we assumed the platform can detect disturbances, it can immediately generate the resistance force without waiting for changes in the HFS. This lets the platform track the trajectory with minimal position error. Subsequently, the HFS adjusts its range to include the neighboring forces around the new equilibrium force (resistance force for the disturbance), implying the platform can address additional force changes. The same adjustment occurs when the disturbance disappears.

